%% file: root.tex
\documentclass[letterpaper,10pt,conference]{ieeeconf}

\IEEEoverridecommandlockouts

\usepackage[export]{adjustbox}
\usepackage{amsmath}
\usepackage{amssymb}
\usepackage{bm}
\usepackage{booktabs}
\usepackage{glossaries}
\usepackage{graphicx}
\usepackage[bookmarks=true]{hyperref}
\usepackage{multirow}
\usepackage[binary-units=true]{siunitx}
\usepackage{subcaption}
\usepackage{times}

\include{acronyms}

\graphicspath{{figures/}}

\hypersetup{hidelinks}

\hypersetup{
    pdfauthor   = {Henrique Ferrolho},%
    pdftitle    = {Optimizing Dynamic Trajectories for Robustness to Disturbances Using Polytopic Projections},%
    pdfsubject  = {Robotics},%
    pdfkeywords = {Robots, Trajectory Optimization, Robustness, Polytopes}%
}

\DeclareMathOperator*{\argmax}{\operatornamewithlimits{\arg\!\max}}
\DeclareMathOperator*{\argmin}{\operatornamewithlimits{\arg\!\min}}

\newcommand\refFigure[1]{Fig.~\ref{#1}}
\newcommand\refSection[1]{Section~\ref{#1}}
\newcommand\refTable[1]{TABLE~\ref{#1}}

\newcommand\citet[1]{\cite{#1}}

\title{\LARGE\bf
Optimizing Dynamic Trajectories for Robustness \\ to Disturbances Using Polytopic Projections
}

\author{
    Henrique Ferrolho, Wolfgang Merkt, Vladimir Ivan, Wouter Wolfslag, Sethu Vijayakumar%
    \thanks{
        This research is supported by the EPSRC UK RAI Hub for Offshore Robotics for Certification of Assets (ORCA, grant reference EP/R026173/1), EU H2020 project Memory of Motion (MEMMO, project ID: 780684), and the EPSRC Centre for Doctoral Training in Robotics and Autonomous Systems (EPSRC, grant reference EP/L016834/1).
    }%
    \thanks{
        All authors are with the School of Informatics, University of Edinburgh, United Kingdom.
        Wolfgang Merkt is also with the Oxford Robotics Institute, University of Oxford, United Kingdom.
    }%
    \thanks{
        Email: \texttt{\href{mailto:henrique.ferrolho@ed.ac.uk}{henrique.ferrolho@ed.ac.uk}}.
    }%
}

\begin{document}
\bstctlcite{IEEEexample:BSTcontrol}

\maketitle
\thispagestyle{empty}
\pagestyle{empty}

\begin{abstract}
    This paper focuses on robustness to disturbance forces and uncertain payloads.
    We present a novel formulation to optimize the robustness of dynamic trajectories.
    A straightforward transcription of this formulation into a nonlinear programming problem is not tractable for state-of-the-art solvers,
    but it is possible to overcome this complication by exploiting the structure induced by the kinematics of the robot.
    The non-trivial transcription proposed allows trajectory optimization frameworks to converge to highly robust dynamic solutions.
    We demonstrate the results of our approach using a quadruped robot equipped with a manipulator.
\end{abstract}

\section{Introduction}
\label{sec:introduction}

When an external force is applied to a legged robot with a manipulator it may cause the robot to slip, or to fail to track a path with its end-effector.
Similarly, the performance degrades when the robot poorly estimates how slippery the ground is or how heavy is its payload.
In either case the motion fails because completing the task while compensating for the external force requires the robot to either command more torque to its actuators than they are capable of delivering, to produce unrealistic contact forces, or both.
These limitations impose constraints that the robot motion has to satisfy.
Therefore, one way to look at robustness is to define it as some metric of distance to these constraints, for instance, as the force the robot can compensate for before violating the motion constraints.
This kind of robustness could be optimized over by the robot controller, however, considering robustness during motion planning would allow us to avoid difficult-to-execute motions altogether.

We tackle the problem of robustness against external perturbations and unmodeled payloads for complex legged robots with manipulation capabilities.
We focus on increasing robustness at the planning stage to provide any tracking controller, including robust control schemes, with greater margins of control authority.
In previous work \cite{wolfslag2020optimisation}, we used the \gls{SUF} applied at some link of the robot as a robustness metric for improving single configurations via convex conic optimization.
In this work, we propose a novel formulation to make the computation more tractable and versatile, allowing us to consider the optimization of entire trajectories with nonlinear dynamics.
Our new computational framework enables us to combine \gls{TO} with the \gls{SUF} metric to produce highly robust and dynamic trajectories.

\begin{figure}[t]
    \centering
    \includegraphics[width=\linewidth]{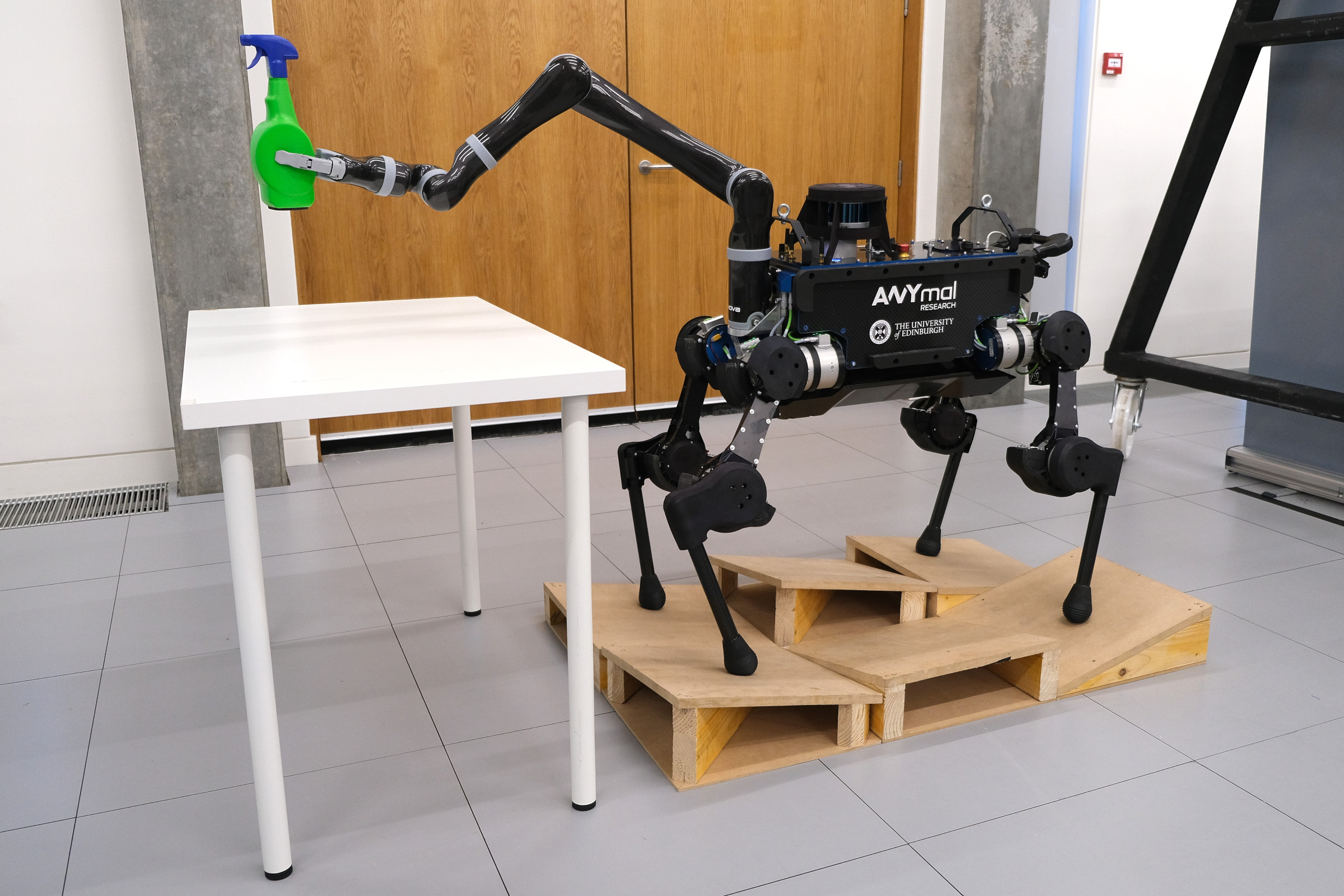}
    \caption{
        A legged loco-manipulation system: ANYmal~\cite{hutter2017anymal} is a fully torque-controlled quadruped robot.
        We equipped it with a Kinova Jaco~\cite{campeau2019kinova} robot arm.
        An accompanying video is available at \texttt{\url{https://youtu.be/vDesP7IpThw}}.
    }\label{figure:cover}
\end{figure}

\subsection{Related Work}

\citet{bellicoso2019alma} presented a motion planning and control framework for a platform similar to ours (see \refFigure{figure:cover}).
The authors demonstrated successful execution of tasks such as opening a door and carrying a box alongside a human.
The authors addressed robustness to external disturbances with an inverse dynamics-based whole-body controller and by re-planning locomotion continuously in a receding-horizon fashion.
However, contrary to our approach, they did not take into account robustness explicitly at the planning-level.

\citet{delprete2016robustness} proposed a solution to improve the robustness to joint-torque tracking errors at the control stage.
The authors modeled deterministic and stochastic uncertainties in joint torques within their control framework optimization.
Their idea is similar to what we present in this paper, but we maximize the upper-bound force magnitude the system can withstand from any possible direction---and we do this during planning.

The authors of \citet{xin2018modelbased} included external forces estimation directly into their hierarchical controller.
Their objective was to minimize actuator torques while enforcing constraints for the contact forces.
However, contrary to our work, they did not enforce actuator limitations explicitly.

Modeling the capabilities of the system explicitly using polytopes has recently become more popular than using simplified metrics for robustness.
In \citet{caron2015leveraging}, the authors derived the equations of a so-called \gls{GIWC}.
It is a feasible region used as a general stability criterion.
This representation is very efficient for testing robust static equilibrium of a legged robot, but it fails to take into account any actuation limits.
\citet{orsolino2018application} proposed to incorporate the properties of \cite{caron2015leveraging} with system torque limits.
They use the resulting polytopes to optimize the \gls{com} trajectory in the $xy$-plane for the base-transfer motion of a quadruped.
Despite the reduced size of this problem, the technique used to compute polytopes was prohibitively expensive, and as a workaround they computed polytopes once at the beginning and used them as an approximation for the remaining motion.

We have followed this line of research in our previous work \cite{ferrolho2019residual} and we proposed a force polytope representation considering system dynamics: the \emph{residual force polytope}.
The polytope is computed from the forces and torques remaining after accounting for Coriolis, centrifugal, and gravity terms, as well as nominal motion feed-forward torques.

All the polytope calculations proposed in the literature \cite{yoshikawa1985manipulability,orsolino2018application,ferrolho2019residual} require a significant amount of computation time.
In general, deriving an explicit description of a projected polytope is NP-hard \cite{tiwary2008computing}.
As a result, prior work using polytopes in trajectory optimization, e.g., \citet{orsolino2018application}, resorted to the approximation of fixing the polytope for an entire trajectory.

\citet{zhen2018computing} recently formulated a computationally tractable approach for finding maximally sized convex bodies inscribed in a projected polytope.
Their scheme does not require an explicit description of the projection and works by combining Fourier-Motzkin elimination with techniques from adjustable robust optimization.
The scheme was adapted for robustness computations in robotics in \cite{wolfslag2020optimisation}, where the \gls{SUF} were estimated for static configurations.
However, despite an improvement over exact computation, due to the computational complexity of their formulation it was not previously possible to consider trajectory optimization of full system dynamics and maximization of robustness based on dynamic polytopes at the same time.
We further adapt the technique of \citet{zhen2018computing} to reformulate the problem of computing the \gls{SUF}.
The resulting reformulation allows to consider trajectory optimization and robustness maximization in a bilevel optimization setting.

Bilevel optimizations are mathematical programs that include the solution to other programs in their constraints or objectives.
They are common in robustness settings, and have been used for robust control of robots.
\cite{landry2019differentiable} optimized trajectories with full dynamics for robustness as a bilevel problem; it is particularly related to our work, but with some key differences:
they considered robustness to noise and dealt with a fixed base manipulator---both differences allowed for simplifications in their optimization problem.

\subsection{Statement of Contributions}
We present a \gls{TO} framework capable of planning robust and dynamic manipulation tasks for legged robots, such as the one shown in \refFigure{figure:cover}.
Our main contributions are:
\begin{enumerate}
    \item Proposal of a novel solution to a bilevel optimization problem that marries dynamic trajectory optimization with maximization of robustness against disturbances.
    \item Explanation of the non-trivial transcription and reformulation required to make this problem tractable for a \gls{NLP} solver.
    \item Comparison of our method's results against a traditional optimization objective across different scenarios. %
    \item Validation of the planned motions using both full-physics simulation and real-life hardware experiments.
\end{enumerate}
We will also make our framework/implementation available upon acceptance/publication of this paper.

\section{Trajectory Optimization}
\label{sec:trajectory_optimization}

Trajectory Optimization (TO) is a well-known and powerful framework for planning locally-optimal trajectories of dynamic systems such as legged robots subject to constraints.
\gls{TO} falls under the broader category of optimal control problems.
In general, \gls{TO} aims to design a finite-time control trajectory as a function of time, $u(t)$, which drives the system from an initial state $x(t_I)$ towards a final state $x(t_F)$, and given the system dynamics $\dot{x} = f(x, u)$ which must be satisfied over the entire interval $t_I \le t \le t_F$.
Optimal control problems can be solved using dynamic programming or by means of transcription (see \citet{betts2010practical}).

In this work, we employ a technique called direct transcription because it readily handles strict constraints on states and controls.
Such constraints take a key role in computing the \gls{SUF}.
The main alternative, \gls{DDP}, offers faster computation and provides a linear controller next to the optimized trajectory.
However, handling constraints with \gls{DDP} is a challenging subject of research~\cite{giftthaler2018family,mastalli2020crocoddyl}.
This currently makes \gls{DDP} less applicable to our case.
A second alternative, shooting methods, have been reported to result in slower computations and higher susceptibility to local optima~\cite{betts2010practical}.
Using direct transcription, we formulate the continuous optimization problem by explicitly discretizing the system state and control trajectories.
This method results in the formulation of a large and sparse \gls{NLP} problem~\cite{betts2010practical}.
The resulting constrained nonlinear optimization problem can then be solved using a sparse, large-scale nonlinear programming solver such as Knitro~\cite{byrd2006knitro}.

\section{Model Formulation}
\label{sec:model_formulation}

The model of a legged robot can be formulated as a free-floating base $B$ to which limbs are attached.
For the specific case of the robot shown in \refFigure{figure:cover}, the kinematic tree stemming from the base branches into four legs and one robotic manipulator with six \gls{DoF}.
The motion of the system can be described \gls{wrt} a fixed inertial frame $I$.
Let us express the position of the base \gls{wrt} the inertial frame, expressed in the inertial frame, as ${}_I\bm{r}_{IB} \in \mathbb{R}^3$.
Let $\bm{q}_{IB} \in \mathbb{H}$ be a Hamiltonian unit quaternion defining the orientation of the free-floating base \gls{wrt} the inertial frame,
and let $\bm{\psi}_{IB} \in \overline{\mathbb{R}}^3$ be the \gls{MRP}~\cite{gormley1945stereographic,terzakis2018modified} of the unit quaternion $\bm{q}_{IB}$.\footnote{%
    $\overline{\mathbb{R}} = \mathbb{R} \cup \left \{ -\infty, +\infty \right \}$ is the \textit{affinely extended set of real numbers}.
    We use the same notation as \citet{terzakis2018modified}.
}
We use $\bm{\psi}_{IB}$ to parameterize the orientation of the free-floating base.\footnote{%
    The \gls{MRP} encode a 3D rotation with the stereographic projection of a Hamiltonian unit quaternion.
    The derivatives of the rotation matrix \gls{wrt} the \gls{MRP} parameters are rational functions, making this representation a particularly good choice for purposes of differentiation and optimization.
}
The joint angles describing the configuration of the 6-\gls{DoF} arm and the four 3-\gls{DoF} legs are stacked in a vector $\bm{q}_j \in \mathbb{R}^{n_j}$, where $n_j = 18$.
The generalized coordinates vector $\bm{q}$ and the generalized velocities vector $\bm{v}$ of this floating-base system may therefore be written as
\begin{equation}
    \bm{q} = \begin{bmatrix} {}_I\bm{r}_{IB} \\ \bm{\psi}_{IB} \\ \bm{q}_j \end{bmatrix} \in \mathbb{R}^3 \times \overline{\mathbb{R}}^3 \times \mathbb{R}^{n_j}, \quad
    \bm{v} = \begin{bmatrix} \bm{\nu}_B \\ \bm{\dot{q}}_j \end{bmatrix} \in \mathbb{R}^{n_v},
\end{equation}
where $n_v = 6 + n_j$ and the \textit{twist} $\bm{\nu}_B = \left [ {}_I\bm{v}_{B} \quad {}_B\bm{\omega}_{IB} \right ] \in \mathbb{R}^6$ encodes the linear and angular velocities of the base $B$ w.r.t. the inertial frame expressed in the $I$ and $B$ frames, respectively.
The equations of motion of a floating base system that interacts with the environment are written as
\begin{equation}
    \bm{M}(\bm{q})\bm{\dot{v}} + \bm{h}(\bm{q}, \bm{v}) = \bm{S}^\top \bm{\tau} + \bm{J}_s^\top(\bm{q}) \bm{\lambda} + \bm{J}_e^\top(\bm{q}) \bm{f},
    \label{equation:equations_of_motion}
\end{equation}
where $\bm{M}(\bm{q}) \in \mathbb{R}^{n_v \times n_v}$ is the mass matrix and $\bm{h}(\bm{q}, \bm{v}) \in \mathbb{R}^{n_v}$ is the vector of Coriolis, centrifugal, and gravity terms.
The selection matrix $\bm{S} = [\bm{0}_{n_\tau \times (n_v - n_\tau)} \quad \mathbb{I}_{n_\tau \times n_\tau}]$ selects which \gls{DoF} are actuated.
Here, $n_\tau = n_j$ as all limb joints are actuated.
The vector of ground-feet contact forces and contact torques $\bm{\lambda}$ is mapped to joint-space torques through the support Jacobian $J_s \in \mathbb{R}^{n_s \times n_v}$, which is obtained by stacking the Jacobians which relate generalized velocities to limb end-effector motion as $\bm{J}_s = [\bm{J}_{C_1}^\top \quad \cdots \quad \bm{J}_{C_{n_c}}^\top]^\top$, with $n_c$ being the number of limbs in contact and $n_s$ the total dimensionality of all contact wrenches.
We assume ANYmal has point-feet and thus we only model linear contact forces at the feet.
Finally, $\bm{f}$ represents any external force applied to the end-effector.
This force may be the result of a push or some unpredicted disturbance.
In a nominal scenario, this force is zero, i.e., $\bm{f} = \bm{0}$.
The Jacobian $J_e \in \mathbb{R}^{3 \times n_v}$ is used to map a linear force $\bm{f}$ applied at the end-effector to joint-space torques. %

\section{Problem Formulation}
\label{sec:problem_formulation}

We transcribe the continuous optimization problem by explicitly discretizing the system state and the control trajectory using a \textit{direct transcription} technique.
We divide the trajectory into $N$ equally spaced segments or intervals
\begin{equation}
    t_I = t_1 < t_2 < \dots < t_M = t_F,
\end{equation}
where the points are referred to as \textit{mesh points}.\footnote{Some authors also refer to these mesh points as \textit{nodes}, \textit{knots}, \textit{way points}, or \textit{grid points}.}
The number of mesh points is given by $M = N + 1$.
Henceforth, we use $x_k \equiv x(t_k)$ and $u_k \equiv u(t_k)$ to indicate the value of the state and control variables, respectively, at mesh point $k$.
We treat the values of $x_k$ and $u_k$ as a set of \gls{NLP} variables, and we finally formulate the general \gls{TO} problem as:
\begin{equation}
    \begin{aligned}
        \argmin_{\bm{\xi}} \qquad   & g_M(x_M) + \sum_{k=1}^{M-1} g(x_k, u_k) \\
        \mathrm{subject\ to} \qquad & x_{k+1} = x_k + h \, f(x_k, u_k)     \\
                                    & x_k \in \mathcal{X}                     \\
                                    & u_k \in \mathcal{U}
    \end{aligned}
    \label{equation:nlp}
\end{equation}
where $g(\cdot, \cdot)$ and $g_M(\cdot)$ form an optional cost function,
$h = (t_F - t_I) / N$ is a fixed \textit{integration step size},
and $\mathcal{X}$ and $\mathcal{U}$ are the sets of feasible states and inputs, respectively.
We use the explicit Euler method to integrate the differential equations of the system dynamics, but other \textit{K}-stage Runge-Kutta schemes could be used, e.g., the Trapezoidal method (implicit, $K = 2$) or the Hermite-Simpson method (implicit, $K = 3$).

\subsection{Parameterization}
Similarly to \citet{posa2014direct}, we directly optimize over the space of feasible states, control inputs, and constraint forces,
i.e., for each discretized mesh point $k$, the vectors of generalized coordinates $\bm{q}_k$, generalized velocities $\bm{v}_k$, control inputs $\bm{\tau}_k$, and contact forces $\bm{\lambda}_k$ form the vector of decision variables $\bm{\xi}_k$.
The entire vector of \gls{NLP} decision variables is:
\begin{align}
    \bm{\xi} \triangleq \{ \bm{q}_1, \bm{v}_1, \bm{\tau}_1, \bm{\lambda}_1, \cdots, \bm{q}_{N}, \bm{v}_{N}, \bm{\tau}_{N}, \bm{\lambda}_{N}, \bm{q}_M, \bm{v}_M \}.\footnotemark
\end{align}
Methods that treat contact forces as optimization variables are referred to as planning ``through contact''.
This approach increases the number of decision variables, but the problem becomes better conditioned \cite{benoit2019bilevel}.
\footnotetext{Notice that $\bm{\tau}_M$ and $\bm{\lambda}_M$ (i.e., the control and contact forces at the final state) are not required, and thus not part of $\bm{\xi}$.}

\subsection{Objectives}
\label{subsection:objectives}
We consider three different optimization objectives $\mathcal{G}_1$--$\mathcal{G}_3$.
The first objective corresponds to the \textit{feasibility problem}, i.e., a problem with constraints but without any cost to minimize.

The second objective $\mathcal{G}_2$ achieves the minimization of actuator torques and is defined as
\begin{equation}
    \mathcal{G}_2 : \quad \argmin_{\bm{\xi}} \quad \sum_{k=1}^{M-1} \, \bm{\tau}_k^\top \bm{\tau}_k.
    \label{equation:objective_min_tau}
\end{equation}

Finally, objective $\mathcal{G}_3$ corresponds to the maximization of the \gls{SUF} at the end-effector.
$\mathcal{G}_3$ involves a problem reformulation which is explained in \refSection{subsection:maximum_volume_inscribed_ball_of_a_polytopic_projection}---the main contribution presented in this manuscript.

\subsection{Constraints}
We now analyze the constraints formulated in the \gls{NLP} in detail.
\refTable{table:constraints_summary} shows a summary of these constraints.

\subsubsection{Bounds on Decision Variables}
We constrain the joint positions, velocities, and torques to be within their respective lower and upper bounds with \eqref{equation:bounds_position}--\eqref{equation:bounds_torque}.
\begin{align}
     & \bm{q}_L    \le \bm{q}_k    \le \bm{q}_U    &  & \forall k = 1 : M     \label{equation:bounds_position} \\
     & \bm{v}_L    \le \bm{v}_k    \le \bm{v}_U    &  & \forall k = 2 : M - 1 \label{equation:bounds_velocity} \\
     & \bm{\tau}_L \le \bm{\tau}_k \le \bm{\tau}_U &  & \forall k = 1 : M - 1 \label{equation:bounds_torque}
\end{align}
We further fix the initial and final velocities to zero:
\begin{align}
    \bm{v}_1 = \bm{v}_M = \bm{0}.
\end{align}

\subsubsection{Friction Cones}
Similarly to \citet{caron2015leveraging}, we model friction at the contact points using an \textit{inner linear approximation} with a four-sided friction pyramid.
Consider the set of points $\{C_i\}$ where the robot is in contact with its environment.
Let $\bm{n}_i$ and $\mu_i$ be the unit normal and the friction coefficient of the support region at each contact, respectively.
A point contact remains fixed as long as its contact force $\bm{f}^c_i$ lies inside the linearized friction cone directed by $\bm{n}_i$:
\begin{align}
    | \bm{f}^{c}_i \cdot \bm{t}_i | & \leq (\mu_i / \sqrt{2}) (\bm{f}^{c}_i \cdot \bm{n}_i), \label{equation:friction_cone_1} \\
    | \bm{f}^{c}_i \cdot \bm{b}_i | & \leq (\mu_i / \sqrt{2}) (\bm{f}^{c}_i \cdot \bm{n}_i), \label{equation:friction_cone_2} \\
    \bm{f}^{c}_i \cdot \bm{n}_i     & > 0, \label{equation:friction_cone_3}
\end{align}
where $(\bm{t}_i, \bm{b}_i)$ form the basis of the tangential contact plane such that $(\bm{t}_i, \bm{b}_i, \bm{n}_i)$ is a direct frame.

\subsubsection{System Dynamics}
Using explicit Euler integration, we enforce the nonlinear system dynamics ($f$) with a finite set of \textit{defect} (or \textit{gap}) constraints in our \gls{NLP} formulation:
\begin{equation}
    \begin{bmatrix} \bm{q}_{k+1} \\ \bm{v}_{k+1} \end{bmatrix} - \begin{bmatrix} \bm{q}_{k} \\ \bm{v}_{k} \end{bmatrix} - h \, f \left( \begin{bmatrix} \bm{q}_{k} \\ \bm{v}_{k} \end{bmatrix}, \begin{bmatrix} \bm{\tau}_{k} \\ \bm{\lambda}_{k} \end{bmatrix} \right) = \bm{0}.
\end{equation}

\subsubsection{Stationary Feet}
Let the \textit{forward kinematics} function for a foot-point contact $i$ be given by $f^\mathrm{fk}(\bm{q}, i)$.
\begin{equation}
    f^\mathrm{fk}(\bm{q}_k, i) = \bm{p}_i \qquad \forall i = 1 : 4, \ k = 1 : M
\end{equation}

\subsubsection{Gripper Task}
\begin{figure}[ht]
    \centering
    \includegraphics[width=0.7\linewidth]{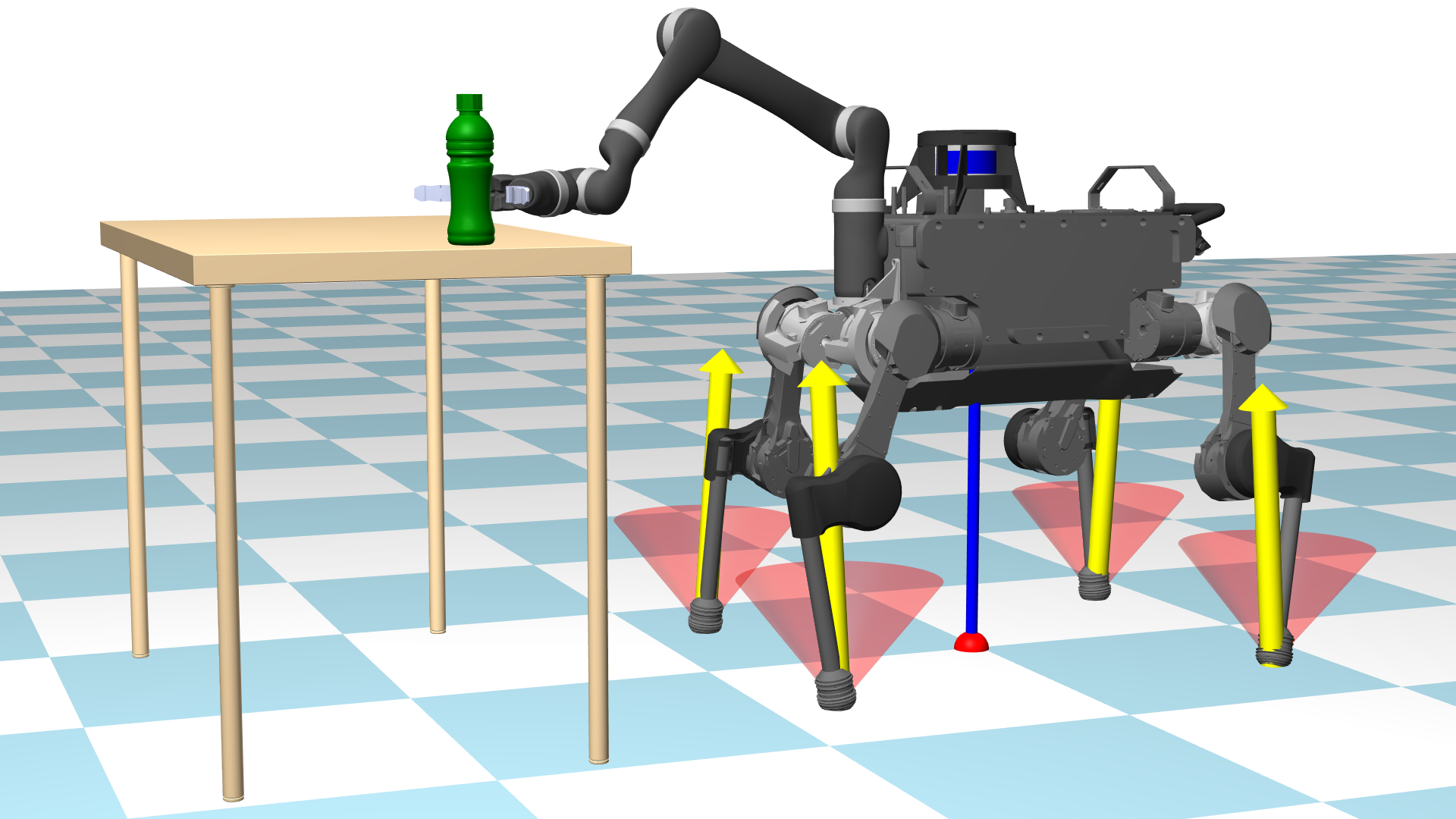}
    \caption{
        The figure shows the robot at the beginning of a pick-and-place task.
        The ground-feet contact forces are shown in yellow and the friction cones are shown in red.
    }\label{figure:robot_picking_bottle}
\end{figure}

The gripper is constrained at the initial and final instants of the trajectory ($k = 1$ and $k = M$).
For both of these mesh points, there exist five constraints: three to constrain the placement of the end-effector, and two for constraining the \textit{pitch} and \textit{roll} describing the orientation of the end-effector.
This enforces a specific location for the pick and placing of a bottle (e.g., see \refFigure{figure:robot_picking_bottle}), as well as the correct orientation of the fingers to embrace it, while leaving the grasp \textit{yaw} as a degree of freedom to the solver.

\begin{table}[h]
    \centering
    \caption{Summary of the formulated \gls{NLP} constraints.}
    \label{table:constraints_summary}
    \begin{tabular}{lcrcr}
        \toprule
        Constraint           &  & Structure &  & Relation   \\
        \midrule
        Bounds on $\bm{\xi}$ &  & Linear    &  & Mixed      \\
        Friction Cones       &  & Linear    &  & Inequality \\
        System Dynamics      &  & Nonlinear &  & Equality   \\
        Stationary Feet      &  & Nonlinear &  & Equality   \\
        Gripper Task         &  & Nonlinear &  & Equality   \\
        \bottomrule
    \end{tabular}
\end{table}

Building on the formulation above, we now want to implement additional terms to model the external disturbances.
We are interested in maximizing the forces applied at the end-effector that the robot can compensate for while still satisfying all the \gls{NLP} constraints from \refTable{table:constraints_summary}.

\section{Robustness to Disturbances}
External forces applied to the robot can cause the robot to slip, lose contact between a foot and the environment, or to fail to track the desired end-effector path.
In each of these cases, the motion fails because the external force causes a violation of one of the motion constrains.
We therefore define robustness as some metric of distance to the constraints.
More specifically, we consider the friction cone constraints on the contact forces and the actuator torque bounds limiting the control commands that can be used to compensate for external forces.
As pointed out by \cite{orsolino2018application}, when transformed into a common reference frame, these constraints form a convex polytope bounding the volume of all admissible external wrenches applied to the robot.
In \cite{ferrolho2019residual} we proposed a geometric method to inscribe a ball into the polytope and use its radius as a metric of robustness.
This approximation is especially useful since the radius of the maximum volume inscribed ball gives a bound for the magnitude of forces from any direction that the system can compensate for without violating the constraints.

\subsection{Maximum Volume Inscribed Ball of a Polytopic Projection}
\label{subsection:maximum_volume_inscribed_ball_of_a_polytopic_projection}

In order to reject a disturbance force, additional motor torques and ground reaction forces are needed.
Our robustness metric, the \gls{SUF}, is the smallest force for which no reaction forces/torques exist that also satisfy friction-cone constraints and motor limitations.
In previous work \cite{ferrolho2019residual}, the \gls{SUF} was computed via a \gls{LP} problem.
The trajectory optimization would have this \gls{LP} problem inside its objective, which is not desirable.
Hence, we propose a new way to compute the \gls{SUF}.
The key idea in this robustness analysis is to approximate the relationship between the disturbance force and reaction forces/torques as affine.
Adapting the results from \cite{zhen2018computing}, we find a practically efficient way to simultaneously optimize the robustness metric and the affine relationship prescribing it.
These results go beyond earlier adaptations in robotics by \cite{wolfslag2020optimisation} because those, like our work relying on \gls{LP}, were not suitable for use in a trajectory optimization setting.\footnote{%
    This is due to the fact that computing derivatives of an LP would require a differentiable solver.
    Solving an LP inside an optimization problem can also lead to higher computational time.
}

Let us define the \textit{extended} torques and ground-feet contact forces as $\bm{\tau}^+$ and $\bm{\lambda}^+$, respectively:
\begin{align}
    \bm{\tau}^+    & = \bm{\tau}    + \bm{K}_{\bm{\tau}}    \bm{f} \label{equation:temp_1}  \\
    \bm{\lambda}^+ & = \bm{\lambda} + \bm{K}_{\bm{\lambda}} \bm{f} \label{equation:temp_2},
\end{align}
where $\bm{\tau}$ and $\bm{\lambda}$ are the \textit{nominal} torques and ground-feet contact forces, $\bm{K}_{\bm{\tau}}$ and $\bm{K}_{\bm{\lambda}}$ are some (instantaneous) gain matrices which map a force expressed in end-effector space to joint-torque space and ground-feet contact space, respectively, and $\bm{f}$ is a potential external force applied at the end-effector.
In a nominal situation, there are no disturbance forces and thus
$\bm{f} = \bm{0}$, $\bm{\tau}^+ = \bm{\tau}$, and $\bm{\lambda}^+ = \bm{\lambda}$.
Assuming no variation in accelerations, replacing $\bm{\tau}$ and $\bm{\lambda}$ in the equations of motion \eqref{equation:equations_of_motion} with the right-hand side of \eqref{equation:temp_1} and \eqref{equation:temp_2} gives:
\begin{equation}
    \bm{0} = \left( \bm{S}^\top \bm{K}_{\bm{\tau}} + \bm{J}_s^\top \bm{K}_{\bm{\lambda}} + \bm{J}_e^\top \right) \bm{f}
    \label{equation:temp_a}
\end{equation}
Alternatively to constraints \eqref{equation:bounds_torque} and \eqref{equation:friction_cone_1}--\eqref{equation:friction_cone_2}, we can represent the actuator torque bounds and friction cones constraints using $\bm{\tau}^+$ and $\bm{\lambda}^+$ as:
\begin{align}
    \bm{A}_{\bm{\tau}}    \bm{\tau}^+    & \le \bm{b}_{\bm{\tau}}    \label{equation:alt_con_a}  \\
    \bm{A}_{\bm{\lambda}} \bm{\lambda}^+ & \le \bm{b}_{\bm{\lambda}} \label{equation:alt_con_b}.
\end{align}
We then substitute \eqref{equation:temp_1}--\eqref{equation:temp_2} into \eqref{equation:alt_con_a}--\eqref{equation:alt_con_b} and for each row $\bm{a}_{\bm{\tau}}$ of matrix $\bm{A}_{\bm{\tau}}$ we write the constraint as:
\begin{equation}
    \bm{a}_{\bm{\tau}}^\top \left( \bm{\tau} + \bm{K}_{\bm{\tau}} \bm{f} \right) \le \bm{b}_{\bm{\tau}} \qquad \forall \left | \bm{f} \right | \le \rho,
    \label{equation:temp_c}
\end{equation}
where $\rho$ is the radius of the maximum volume inscribed ball of a polytopic projection, and it represents the magnitude of the smallest potential disturbance that cannot be directly rejected.
We then define $\bm{f} = \rho \bm{\chi}$, where $\bm{\chi} \in \mathbb{R}^3$ is a vector with unit length, which allows us to find the greatest $\rho$ with:
\begin{equation}
    \left( \max_{\bm{\chi}} \quad \bm{a}_{\bm{\tau}}^\top \left( \bm{\tau} + \bm{K}_{\bm{\tau}} \rho \bm{\chi} \right) \right) \le \bm{b}_{\bm{\tau}}.
    \label{equation:temp_d}
\end{equation}
The objective function of the left-hand side of equation \eqref{equation:temp_d} can be seen as a scalar product of the vectors $\bm{a}_{\bm{\tau}}^\top \bm{K}_{\bm{\tau}} \rho$ and $\bm{\chi}$, which is greatest when these vectors are collinear:
\begin{equation}
    \argmax_{\bm{\chi}} \quad \bm{a}_{\bm{\tau}}^\top \bm{K}_{\bm{\tau}} \rho \bm{\chi} \quad \equiv \quad \frac{\bm{K}_{\bm{\tau}}^\top \bm{a}_{\bm{\tau}}}{\left \| \bm{a}_{\bm{\tau}}^\top \bm{K}_{\bm{\tau}} \right \|}.
    \label{equation:temp_e}
\end{equation}
Simplifying \eqref{equation:temp_d} with the right-hand side of \eqref{equation:temp_e} leads to:
\begin{equation}
    \bm{a}_{\bm{\tau}}^\top \bm{\tau} + \left \| \bm{a}_{\bm{\tau}}^\top \bm{K}_{\bm{\tau}} \right \| \rho \le \bm{b}_{\bm{\tau}}.
    \label{equation:temp_b}
\end{equation}
Equations \eqref{equation:temp_c}--\eqref{equation:temp_b} address the constraints on actuation limits.
We repeat the same process for the ground-feet contact forces to obtain:
\begin{equation}
    \bm{a}_{\bm{\lambda}}^\top \bm{\lambda} + \left \| \bm{a}_{\bm{\lambda}}^\top \bm{K}_{\bm{\lambda}} \right \| \rho \le \bm{b}_{\bm{\lambda}}.
    \label{equation:temp_b_lam}
\end{equation}

Next, we substitute $\overline{\bm{K}}_{\bm{\tau}} = \bm{K}_{\bm{\tau}} \rho$ and $\overline{\bm{K}}_{\bm{\lambda}} = \bm{K}_{\bm{\lambda}} \rho$ into \eqref{equation:temp_a}, \eqref{equation:temp_b} and \eqref{equation:temp_b_lam} and write:
\begin{equation}
    \bm{S}^\top \overline{\bm{K}}_{\bm{\tau}} + \bm{J}_s^\top \overline{\bm{K}}_{\bm{\lambda}} + \bm{J}_e^\top \rho = \bm{0},
    \label{equation:reformulation_constraint_1}
\end{equation}
\begin{equation}
    \bm{a}_{\bm{\tau}}^\top \bm{\tau} + \left \| \bm{a}_{\bm{\tau}}^\top \overline{\bm{K}}_{\bm{\tau}} \right \| \le \bm{b}_{\bm{\tau}},
    \label{equation:reformulation_constraint_2}
\end{equation}
\begin{equation}
    \bm{a}_{\bm{\lambda}}^\top \bm{\lambda} + \left \| \bm{a}_{\bm{\lambda}}^\top \overline{\bm{K}}_{\bm{\lambda}} \right \| \le \bm{b}_{\bm{\lambda}}.
    \label{equation:reformulation_constraint_3}
\end{equation}
This substitution removes the bilinear products between $\bm{K}_{\bm{\tau}}$, $\bm{K}_{\bm{\lambda}}$ and $\rho$ while keeping the equality and inequalities valid.

\subsection{Constraints' Structure Exploitation}

We now extend the problem formulation and transcribe the constraints \eqref{equation:reformulation_constraint_1}--\eqref{equation:reformulation_constraint_3} directly into NLP constraints and we extend the vector of decision variables with $\overline{\bm{K}}_{\bm{\tau}}$, $\overline{\bm{K}}_{\bm{\lambda}}$, and $\rho$.
However, by trying this, one will soon realize we face an NP-hard problem.
Additionally, there is a significant increase in the amount of decision variables, and the dependency of both $\bm{J}_s$ and $\bm{J}_e$ on joint positions means that constraint \eqref{equation:reformulation_constraint_1} is nonlinear and non-convex.
Ultimately, this quickly renders any efforts of a na\"{i}ve transcription ineffective, as the solver would be unable to digest it.

In order to solve this issue, we have to analyze the inherent structure of the problem and its constraints.
Let $\overline{\bm{K}}_{\bm{\tau}}$ be the unknown in constraint \eqref{equation:reformulation_constraint_1}.
Splitting the structure of the constraint as
\begin{equation}
    \begin{bmatrix} \bm{0} \\ \mathbb{I} \end{bmatrix} \overline{\bm{K}}_{\bm{\tau}} = - \begin{bmatrix} \bm{J}_s^{\top_\mathrm{base}} \\ \bm{J}_s^{\top_\mathrm{limbs}} \end{bmatrix} \overline{\bm{K}}_{\bm{\lambda}} - \begin{bmatrix} \bm{J}_e^{\top_\mathrm{base}} \\ \bm{J}_e^{\top_\mathrm{limbs}} \end{bmatrix} \rho
    \label{equation:constraint_splittage}
\end{equation}
highlights that $\overline{\bm{K}}_{\bm{\tau}}$ can be obtained as a function of $\overline{\bm{K}}_{\bm{\lambda}}$ and $\rho$ without performing any inversions.
Doing this satisfies the bottom equality implicitly.
The top part of the equality affecting the floating base still needs to be enforced.

This key-insight into the structure of the constraints allows us to transcribe the problem so that the solver will converge successfully.

\begin{figure*}[ht]
    \centering
    \begin{subfigure}[t]{0.25\linewidth}
        \centering
        \includegraphics[width=0.95\linewidth]{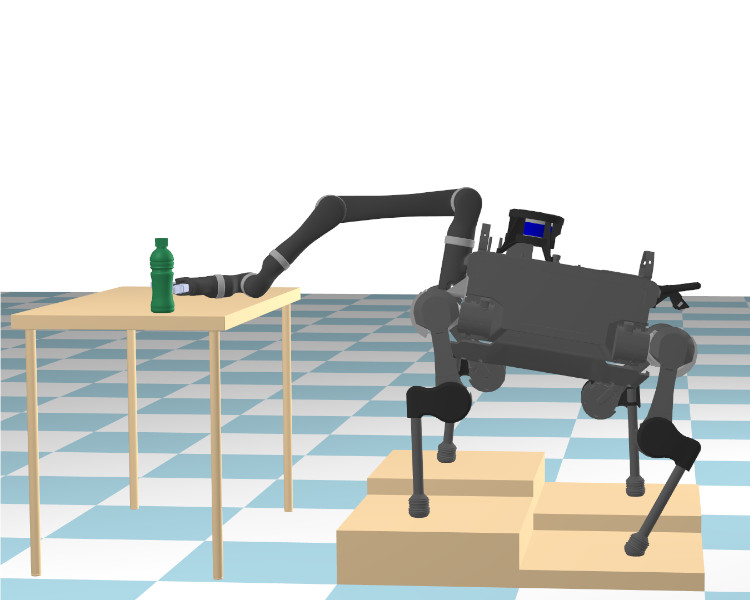}
    \end{subfigure}\hfill%
    \begin{subfigure}[t]{0.25\linewidth}
        \centering
        \includegraphics[width=0.95\linewidth]{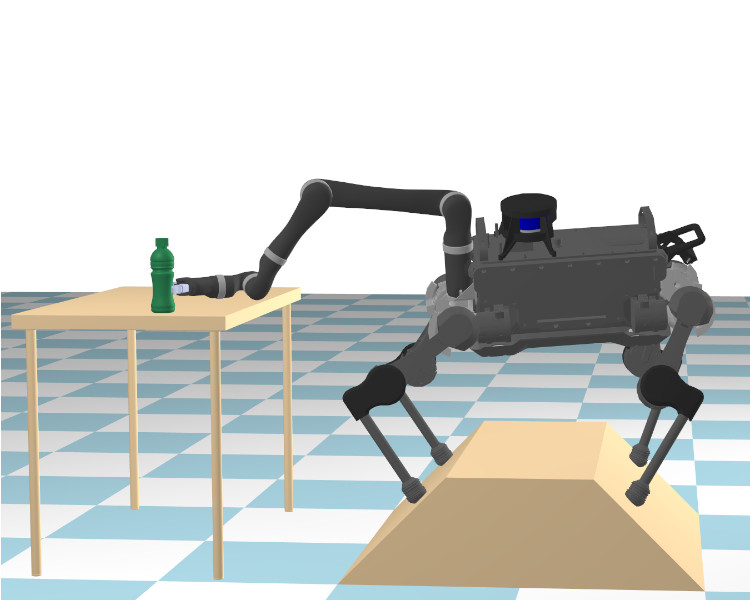}
    \end{subfigure}\hfill%
    \begin{subfigure}[t]{0.25\linewidth}
        \centering
        \includegraphics[width=0.95\linewidth]{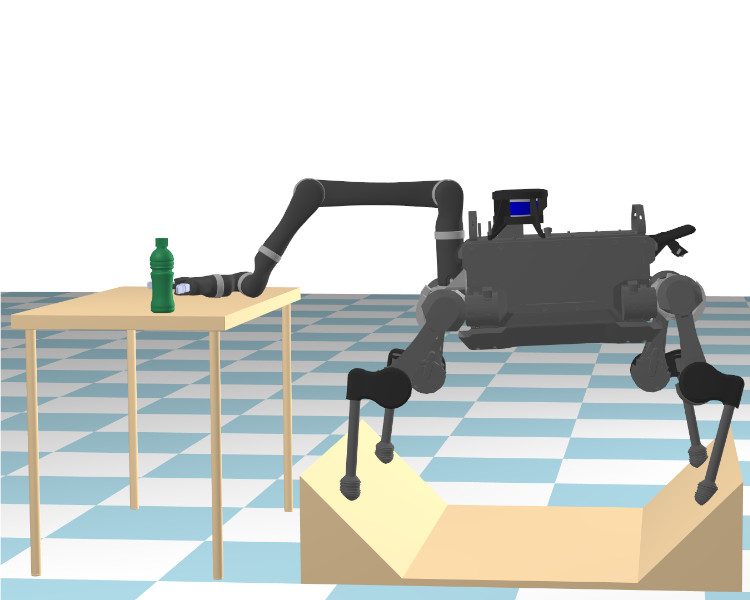}
    \end{subfigure}\hfill%
    \begin{subfigure}[t]{0.25\linewidth}
        \centering
        \includegraphics[width=0.95\linewidth]{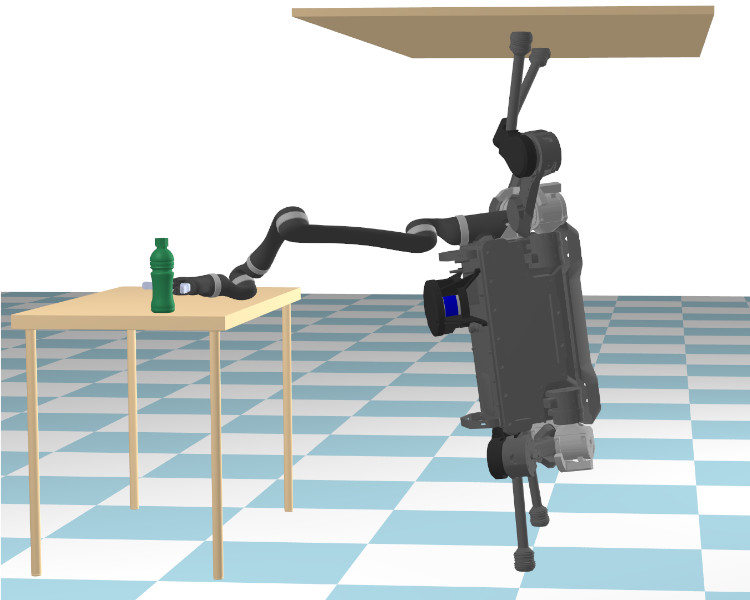}
    \end{subfigure}

    \bigskip
    \begin{subfigure}[t]{0.25\linewidth}
        \centering
        \includegraphics[width=\linewidth]{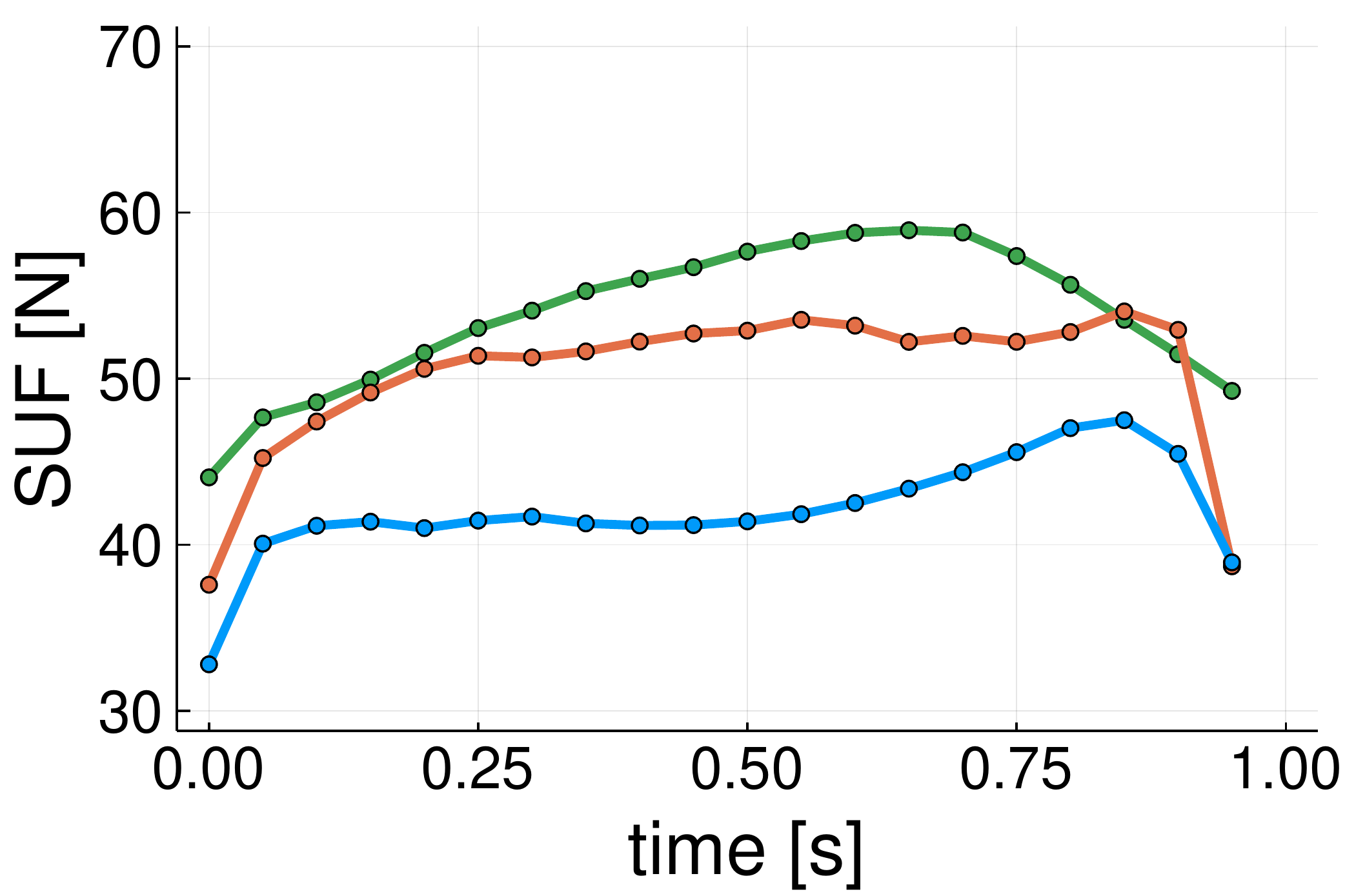}
        \caption{``Steps'' terrain.}
        \label{subfigure:steps}
    \end{subfigure}\hfill%
    \begin{subfigure}[t]{0.25\linewidth}
        \centering
        \includegraphics[width=\linewidth]{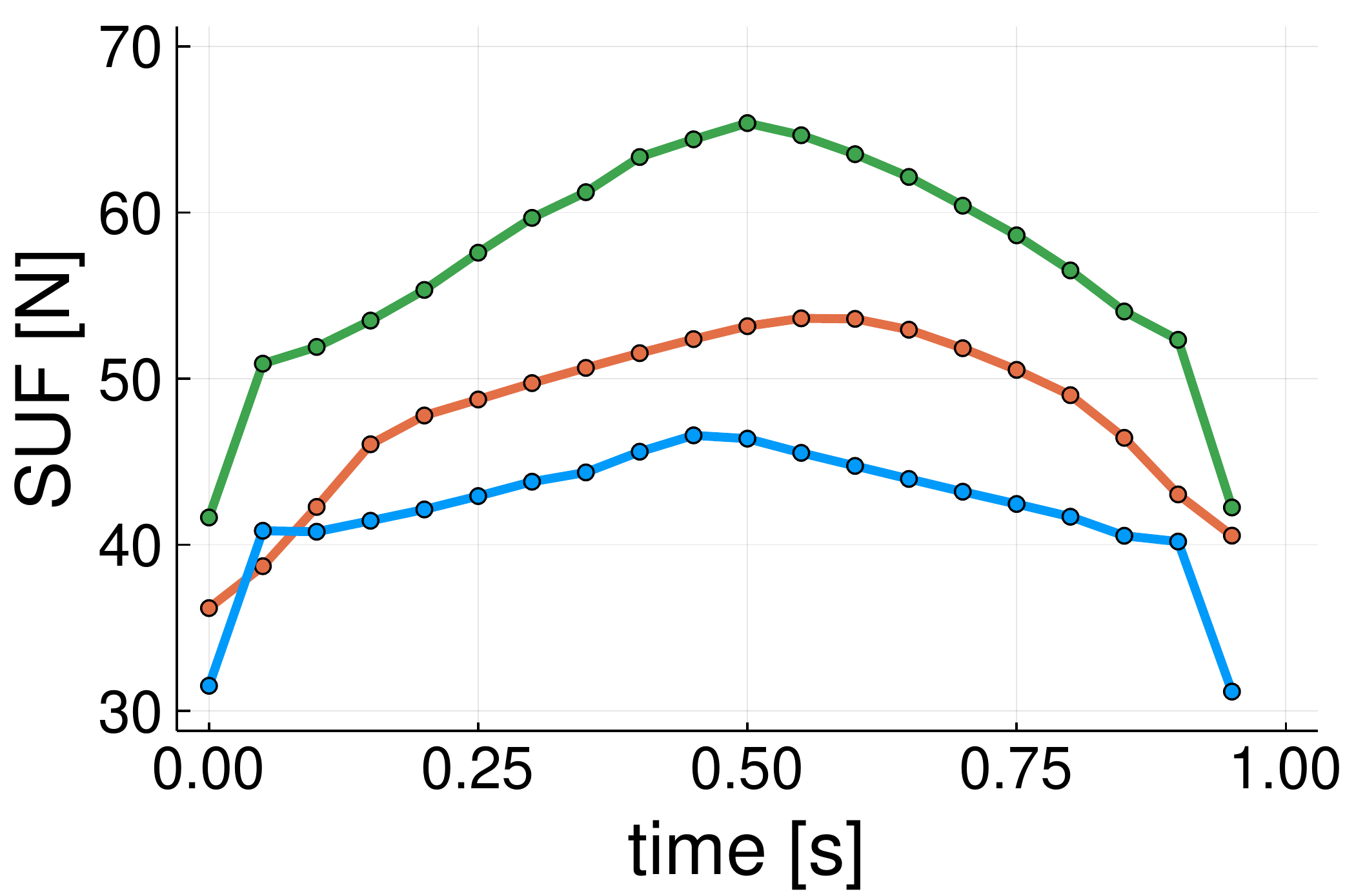}
        \caption{``Ramp'' terrain.}
        \label{subfigure:ramp}
    \end{subfigure}\hfill%
    \begin{subfigure}[t]{0.25\linewidth}
        \centering
        \includegraphics[width=\linewidth]{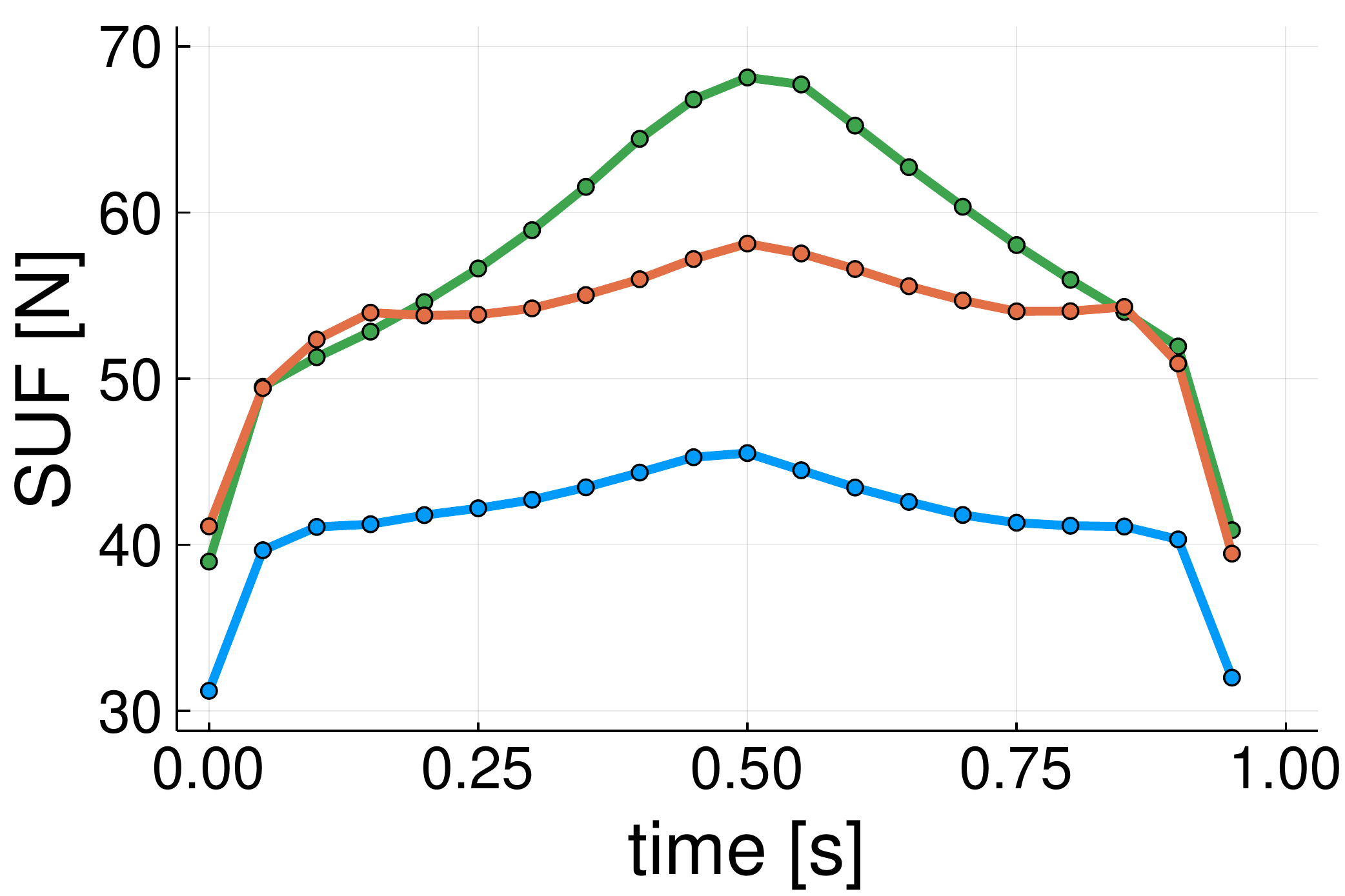}
        \caption{``Valley'' terrain.}
        \label{subfigure:valley}
    \end{subfigure}\hfill%
    \begin{subfigure}[t]{0.25\linewidth}
        \centering
        \includegraphics[width=\linewidth]{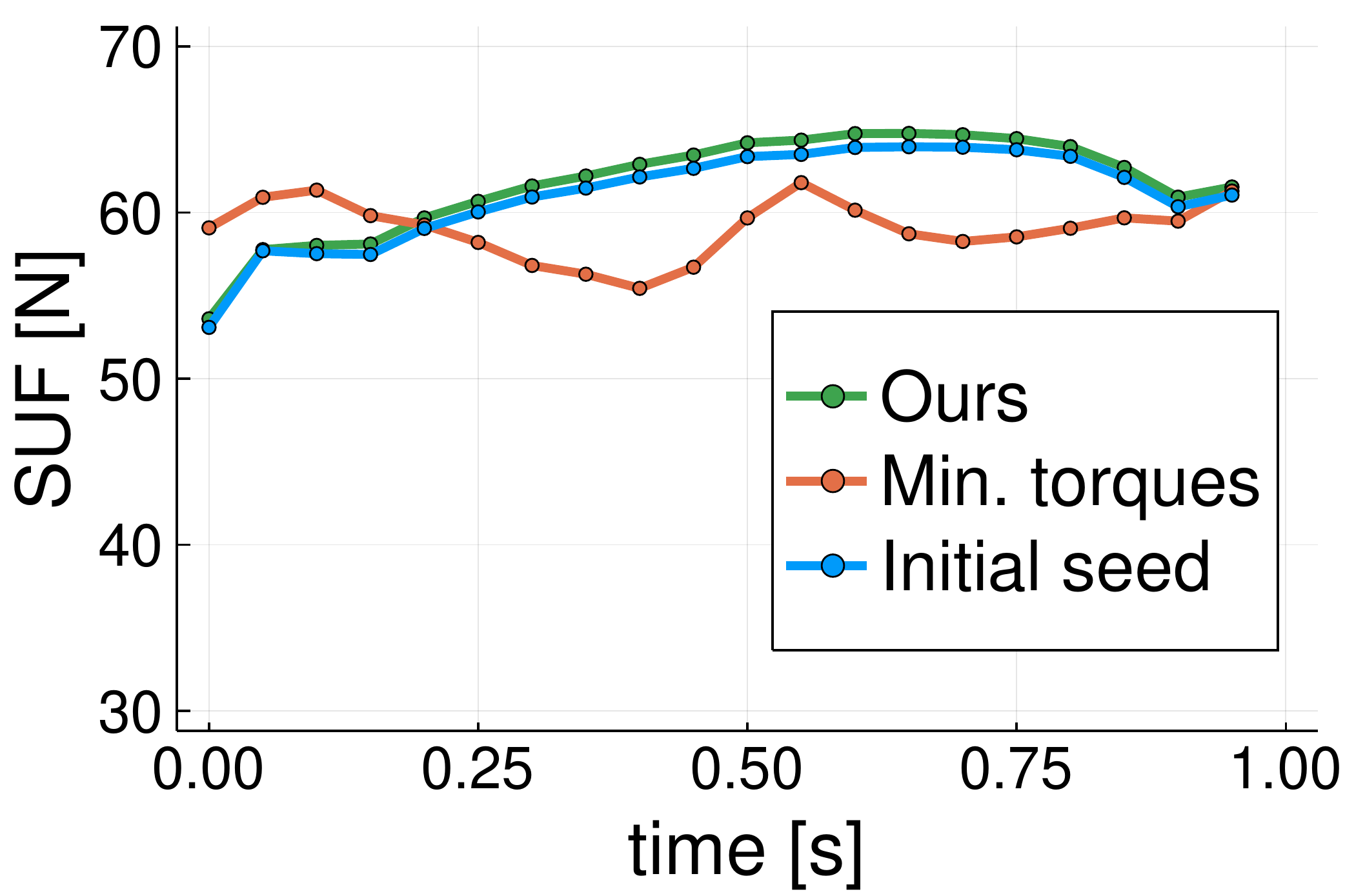}
        \caption{``Handstand'' scenario.}
        \label{subfigure:handstand}
    \end{subfigure}
    \caption{
        We set up a multitude of terrains for testing a pick-and-place task: flat ground, slabs at different heights (a), and inclined supports (b)--(c).
        An extreme scenario where the robot performs a ``handstand'' is shown in (d).
        The plots underneath each scene show the \gls{SUF} (in newtons) for the trajectories computed using different optimization objectives.
    }\label{figure:terrains}
\end{figure*}

\subsection{NLP Reformulation}

\subsubsection{Parameterization}
We extend the previous definition of $\bm{\xi}$ to accommodate for the extra decision variables required.
Recall that ${\overline{\bm{K}}_{\bm{\tau}}}_k$ need not be discretized.
\begin{equation}
    \bm{\xi}^+ \triangleq \bm{\xi} \cup \{ \rho_1, {\overline{\bm{K}}_{\bm{\lambda}}}_1, \cdots, \rho_N, {\overline{\bm{K}}_{\bm{\lambda}}}_N \}.
\end{equation}

\subsubsection{Objective}
$\mathcal{G}_3$ is the sum of all the $\rho_k$ in $\bm{\xi}^+$:
\begin{equation}
    \mathcal{G}_3 : \quad \argmax_{\bm{\xi}^+} \quad \sum_{k=1}^{M-1} \, \rho_k
    \label{equation:objective_max_rho}
\end{equation}

\subsubsection{Constraints}
We bound all the $\rho_k$ in $\bm{\xi}^+$ to $\mathbb{R^+}$ with a linear one-sided inequality:
\begin{equation}
    \rho_k \ge 0 \quad \forall k = 1 : M - 1.
\end{equation}
We enforce the top part of constraint \eqref{equation:constraint_splittage} explicitly:
\begin{equation}
    \bm{J}_s^{\top_\mathrm{base}} \overline{\bm{K}}_{\bm{\lambda}} + \bm{J}_e^{\top_\mathrm{base}} \rho = \bm{0}
    \label{equation:transcribed_constraint_1}
\end{equation}
Finally, \eqref{equation:reformulation_constraint_2} is rewritten as:
\begin{equation}
    \bm{a}_{\bm{\tau}}^\top \bm{\tau} + \left \| \bm{a}_{\bm{\tau}}^\top \left ( - \bm{J}_s^{\top_\mathrm{limbs}} \overline{\bm{K}}_{\bm{\lambda}} - \bm{J}_e^{\top_\mathrm{limbs}} \rho \right ) \right \| \le \bm{b}_{\bm{\tau}},
    \label{equation:transcribed_constraint_2}
\end{equation}

A summary of the constraints added to the \gls{NLP} with the reformulation is shown in \refTable{table:constraints_summary_2}.

\begin{table}[h]
    \centering
    \caption{Summary of the reformulated \gls{NLP} constraints.}
    \label{table:constraints_summary_2}
    \begin{tabular}{lcrcr}
        \toprule
        Constraint                                           &  & Structure &  & Relation   \\
        \midrule
        Bounds on $\rho$                                     &  & Linear    &  & Inequality \\
        Equation \eqref{equation:transcribed_constraint_1}   &  & Nonlinear &  & Equality   \\
        Equation \eqref{equation:transcribed_constraint_2}   &  & Nonlinear &  & Inequality \\
        Equation \eqref{equation:reformulation_constraint_3} &  & Conic     &  & Inequality \\
        \bottomrule
    \end{tabular}
\end{table}

\section{Performance evaluation}
\label{sec:results}

In order to evaluate our work, we compared the robustness of the three objective functions proposed in \refSection{subsection:objectives}: feasibility ($\mathcal{G}_1$), minimum torques ($\mathcal{G}_2$), and maximum \gls{SUF} ($\mathcal{G}_3$).
We ran the comparison across different scenarios for a pick-and-place task.
Furthermore, we benchmarked the times taken to evaluate all \gls{NLP} constraints and the convergence times for problems of different sizes.

\refFigure{figure:terrains} shows four different settings for a pick-and-place task of a bottle on a table.
We set up scenarios with challenging terrain, where the robot stands on steps with different heights or inclined slabs.
The trajectories optimized with our method ($\mathcal{G}_3$) demonstrated greater robustness, as shown in plots (a)--(c) in \refFigure{figure:terrains}.
The initial guess used for $\mathcal{G}_2$ and $\mathcal{G}_3$ was the result of the feasibility problem $\mathcal{G}_1$.

\begin{figure}[b]
    \centering
    \includegraphics[width=\linewidth]{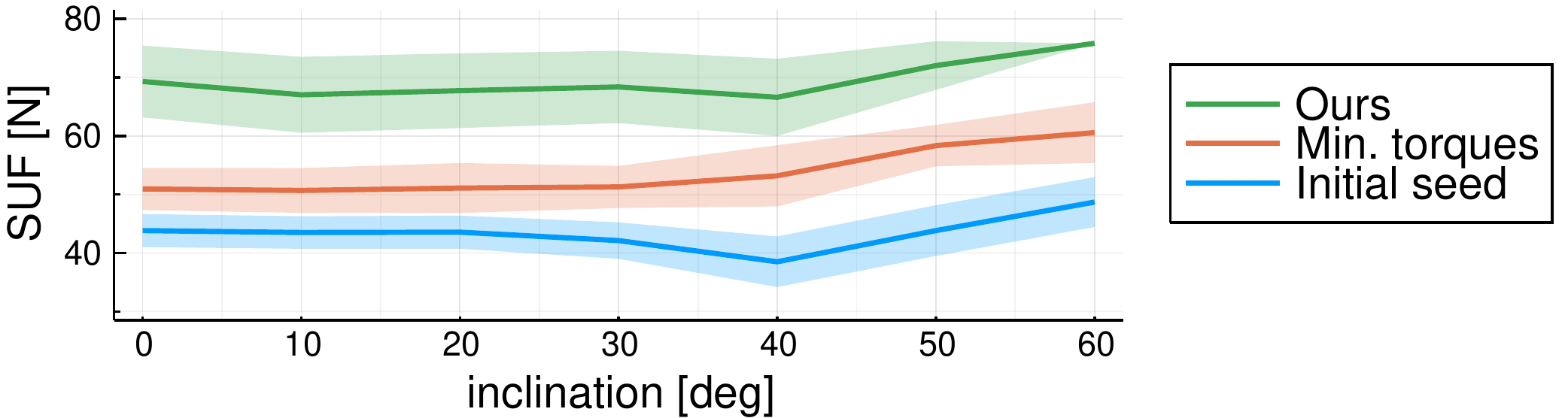}
    \caption{
        Mean and standard deviation of the \gls{SUF} at the end-effector for varying inclinations on the ``ramp'' scenario.
    }\label{figure:ramp_data}
\end{figure}

We also verified the robustness of trajectories for different inclines.
For this, we varied the grade of the slopes in the ``ramp'' scenario (\refFigure{subfigure:ramp}) from \SI{0}{\degree} to \SI{60}{\degree}.
The trajectories computed with our metric consistently showed a greater \gls{SUF} for all the tested slopes (see \refFigure{figure:ramp_data}).

\begin{figure*}[ht]
    \centering
    \begin{subfigure}[t]{0.333\linewidth}
        \includegraphics[width=0.95\linewidth,left]{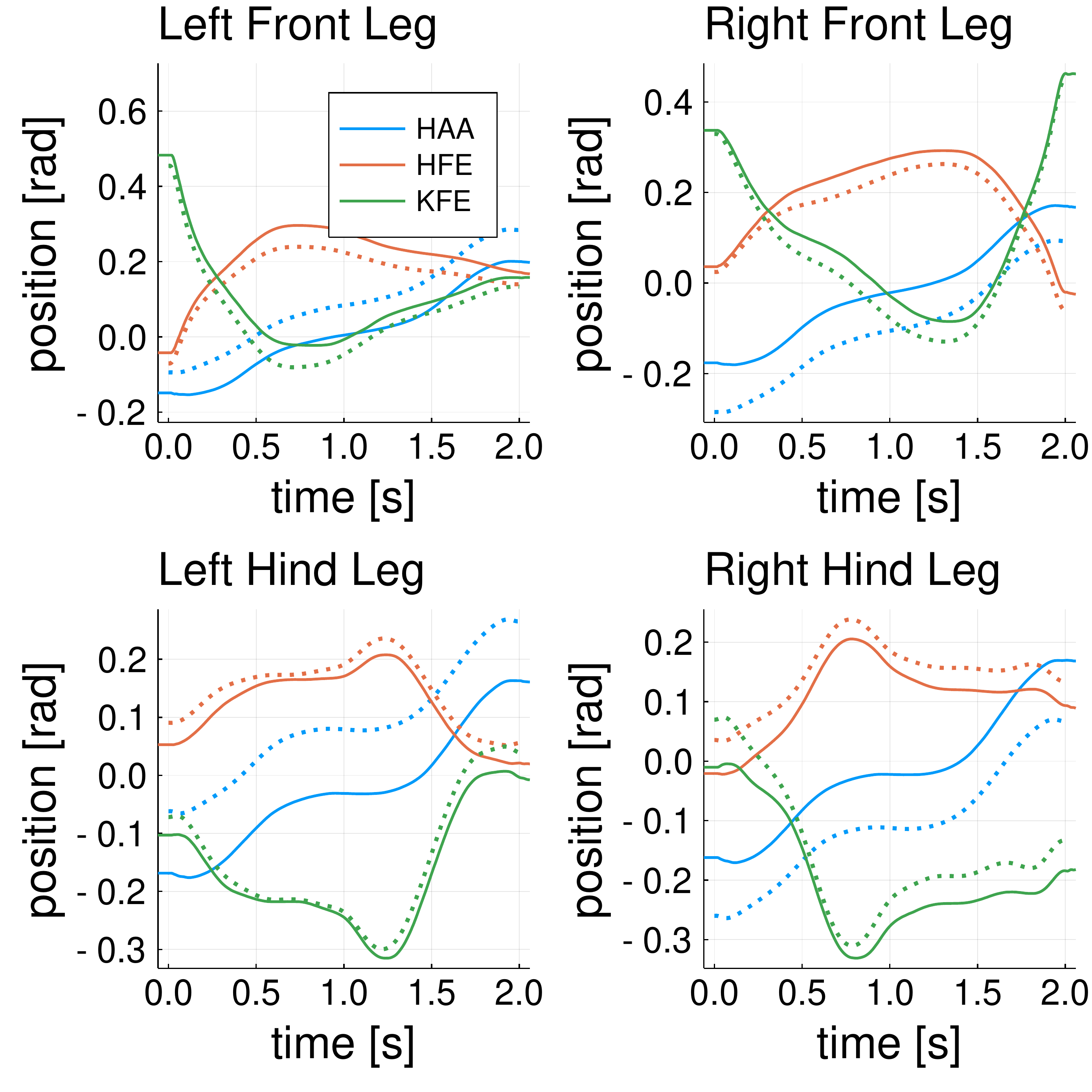}
        \caption{Position}
    \end{subfigure}\hfill%
    \begin{subfigure}[t]{0.333\linewidth}
        \includegraphics[width=0.95\linewidth,center]{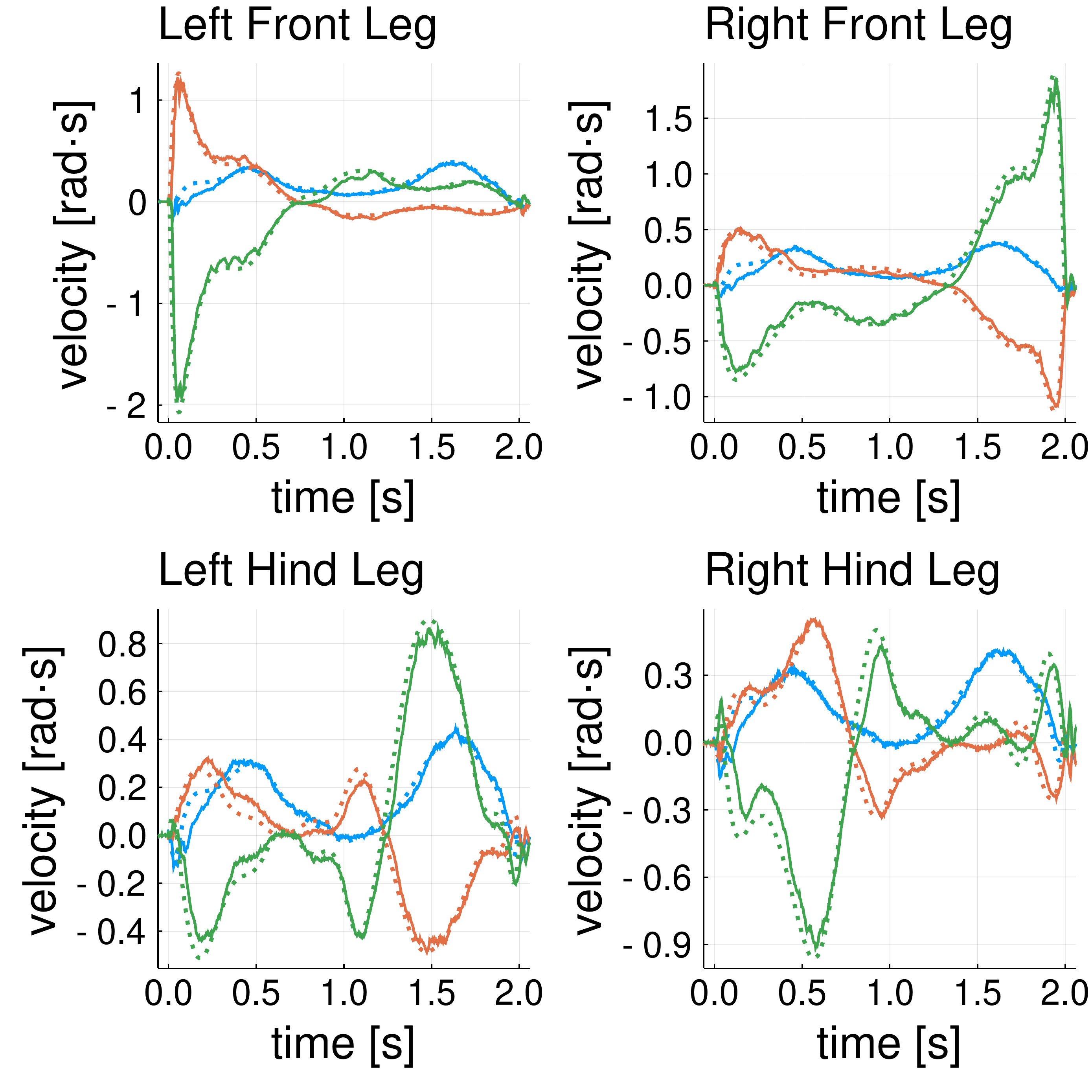}
        \caption{Velocity}
    \end{subfigure}\hfill%
    \begin{subfigure}[t]{0.333\linewidth}
        \includegraphics[width=0.95\linewidth,right]{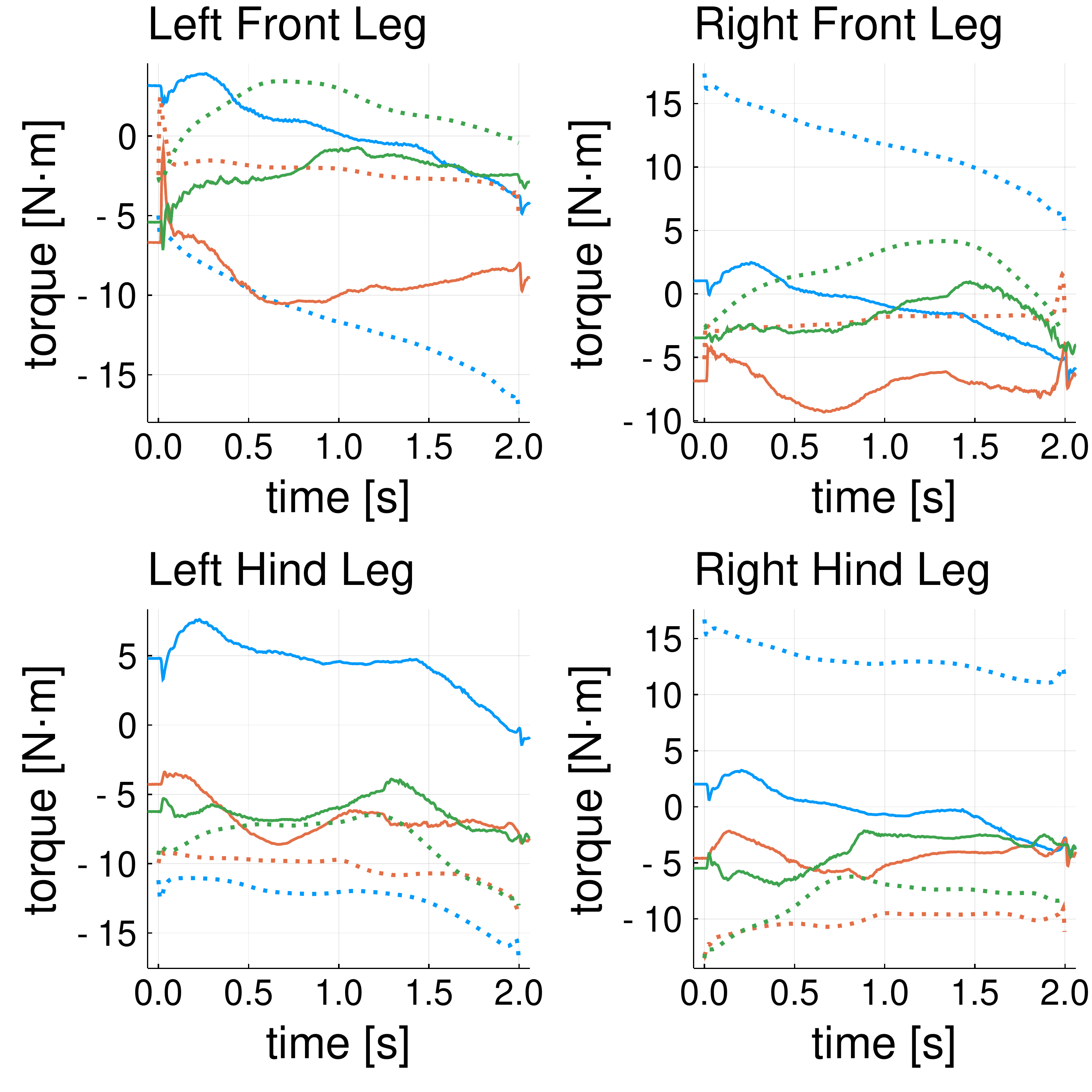}
        \caption{Effort}
    \end{subfigure}
    \caption{
        Joint positions, velocities, and torques of ANYmal for a 2-seconds long trajectory on flat ground.
        The dotted lines correspond to the planned trajectory.
        The solid lines show the data collected by the state estimation on the real robot.
    }\label{figure:rosbag}
\end{figure*}

As an extreme example, we created a scenario where the robot has to perform a ``handstand'', i.e., support its own weight on two of its legs (see \refFigure{subfigure:handstand}) while using the remaining two for keeping its balance.
In this scenario, it is especially important for the robot to press downwards against the floor and upwards against the ceiling to maintain stability.
Because of this, torque minimization ($\mathcal{G}_2$) is not an appropriate objective for this scenario, as confirmed by the degraded \gls{SUF} when compared to the initial seed in plot (d).
On the other hand, our method is able to increase the robustness of the initial seed by a small amount, because it allows to trade off torque expenditure for more stable ground/ceiling-feet contact forces.
We would like to emphasize that the motions in all of the scenarios are within the actual physical capabilities of the robot, even the ``handstand'' scenario.

\begin{table}[t]
    \centering
    \caption{Times taken to evaluate the \gls{NLP} constraints.}
    \label{table:times_fc_ga}
    \begin{tabular}{
        l
        c
        S[table-format=2.2]@{\,\( \pm \)\,}
        S[table-format=3.2]
        S[table-format=4.2]@{\,\( \pm \)\,}
        S[table-format=4.2]}
        \toprule
        Constraint                                        &   &
        \multicolumn{2}{c}{Function (\si{\micro\second})} &
        \multicolumn{2}{c}{Jacobian (\si{\micro\second})}                                          \\
        \midrule
        Gripper Task                                      &   & 9.15  & 25.18  & 14.93   & 2.47    \\
        Stationary Feet                                   &   & 18.29 & 2.07   & 57.67   & 225.53  \\
        System Dynamics                                   &   & 55.07 & 169.64 & 3801.85 & 1353.08 \\
        SUF Constraints                                   &   & 73.76 & 239.97 & 1396.90 & 989.31  \\
        \bottomrule
    \end{tabular}
\end{table}
\refTable{table:times_fc_ga} shows the computation times for function and Jacobian evaluations of the \gls{NLP} problem constraints.
The longest time is spent computing the Jacobian of the system dynamics.
Evaluating the Jacobian of the \gls{SUF}---which is involved when optimizing $\mathcal{G}_3$---takes the second longest time.

\begin{table}[t]
    \centering
    \caption{Convergence times of objectives $\mathcal{G}_1$--$\mathcal{G}_3$ for problems with different size: $11$, $21$, and $41$ mesh points.}
    \label{table:times_convergence}
    \begin{tabular}{c
        c
        S[table-format=1.2]@{\,\( \pm \)\,}S[table-format=1.3]
        S[table-format=3.2]@{\,\( \pm \)\,}S[table-format=2.2]
        S[table-format=4.2]@{\,\( \pm \)\,}S[table-format=2.2]}
        \toprule
        $M$                                                &   &
        \multicolumn{2}{c}{$\mathcal{G}_1$ (\si{\second})} &
        \multicolumn{2}{c}{$\mathcal{G}_2$ (\si{\second})} &
        \multicolumn{2}{c}{$\mathcal{G}_3$ (\si{\second})}                                                       \\
        \midrule
        11                                                 &   & 0.46 & 0.007 & 115.45 & 0.27  & 229.34  & 0.39  \\
        21                                                 &   & 0.74 & 0.009 & 143.48 & 5.56  & 608.09  & 8.04  \\
        41                                                 &   & 1.21 & 0.005 & 835.81 & 15.59 & 1775.23 & 12.85 \\
        \bottomrule
    \end{tabular}
\end{table}
\refTable{table:times_convergence} shows the total time it takes for the solver to converge for problems of different size.
We benchmarked problems with $11$, $21$, and $41$ mesh points (for a 1-second trajectory, this is the equivalent of a discretization at $10$, $20$, and \SI{40}{\hertz}).
Each average and standard deviation are taken from five samples.
$\mathcal{G}_1$ is the fastest to solve as it is a feasibility problem and does not consider any optimality function.
In our tests, the overall best robustness of $\mathcal{G}_3$ (shown in \refFigure{figure:terrains}) also comes with the trade-off of the longest times required until convergence.

All evaluations in this section were carried out in a single-threaded process on an Intel i7-6700K CPU with \SI{4.0}{\giga\hertz} and \SI{32}{\giga\byte} \SI{2133}{\mega\hertz} memory.
The proposed optimization framework has been implemented using Julia~\cite{bezanson2017julia} and the optimization library Knitro~\cite{byrd2006knitro}.
The chosen solving algorithm was the interior-point method\footnotemark\ presented by \citet{waltz2006interior}.
\footnotetext{%
    Interior-point methods (also known as barrier methods) replace the \gls{NLP} problem by a series of barrier subproblems controlled by a barrier parameter.
    They are generally preferable for large-scale problems.
}

\section{Experiments}
\label{sec:experiments}

We conducted hardware experiments on an ANYmal~\cite{hutter2017anymal} quadruped equipped with a Kinova Jaco~\cite{campeau2019kinova} robot arm.
The motion planning is performed a priori and the optimized trajectories are then sent to the controller for playback.

\subsection{Robot Control}
\label{subsec:robot_control}
To execute our whole-body motions, we tracked the joint position with feed-forward velocity and torque.
We updated the references for joint position, joint velocity, and joint torque at \SI{400}{\hertz}.
The decentralized motor controller at every joint closes the loop compensating for friction effects. %
During our experiments, we used $k_p = [150, 150, 100]$ as proportional and $k_d = [0.5, 0.5, 0.45]$ as derivative joint space gains for each leg, respectively.
The arm used Kinova's driver for joint trajectory control.
We synchronized the execution of both controllers.

In order to evaluate how well the real robot tracks motions using our controller, we compared the planned joint states over time with the state estimation data from the robot.
We computed a 2-seconds long trajectory using our framework for a pick-and-place task sampled at \SI{400}{\hertz} and commanded the robot at the same frequency. %
\refFigure{figure:rosbag} shows the plots of the planned trajectory against the data collected during our experiments.
The plots show that joint positions are within acceptable tolerances and joint velocities are tracked well, but joint efforts are significantly different.
This validates that the motions generated by our trajectory optimization are dynamically consistent.
The mismatch in joint efforts is expected due to differences between the real robot and our model, and also due to signal delays.
Additionally, as we executed our trajectory open-loop without re-planning, the errors accumulated.
Nonetheless, the tracking controller can execute the dynamic motions we planned.

\subsection{Description of the Experiments}
\begin{figure}[ht]
    \begin{subfigure}[t]{0.5\linewidth}
        \centering
        \includegraphics[width=0.99\linewidth,left]{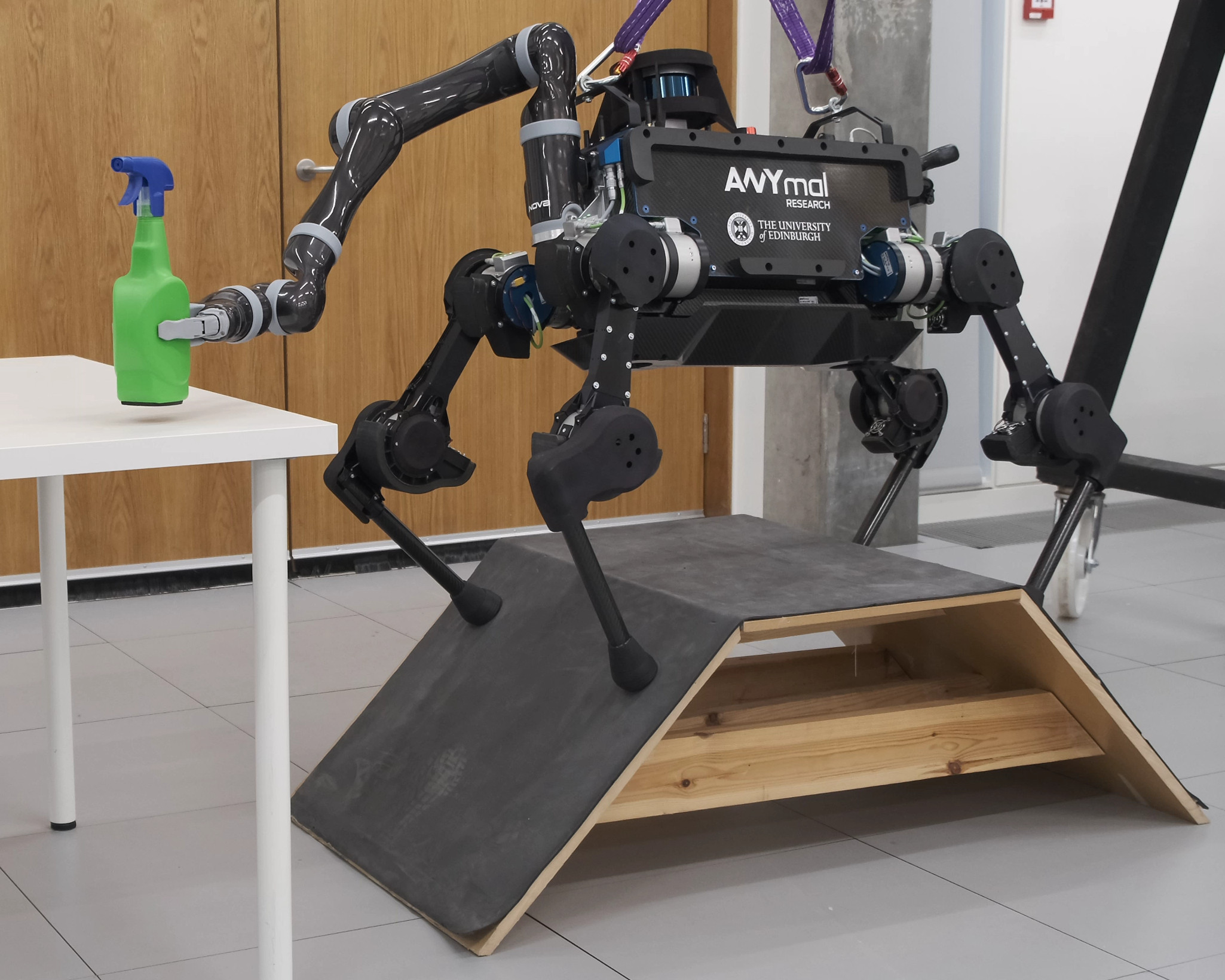}
        \caption{``Ramp'' terrain}
        \label{subfigure:experiment_ramp}
    \end{subfigure}\hfill%
    \begin{subfigure}[t]{0.5\linewidth}
        \centering
        \includegraphics[width=0.99\linewidth,right]{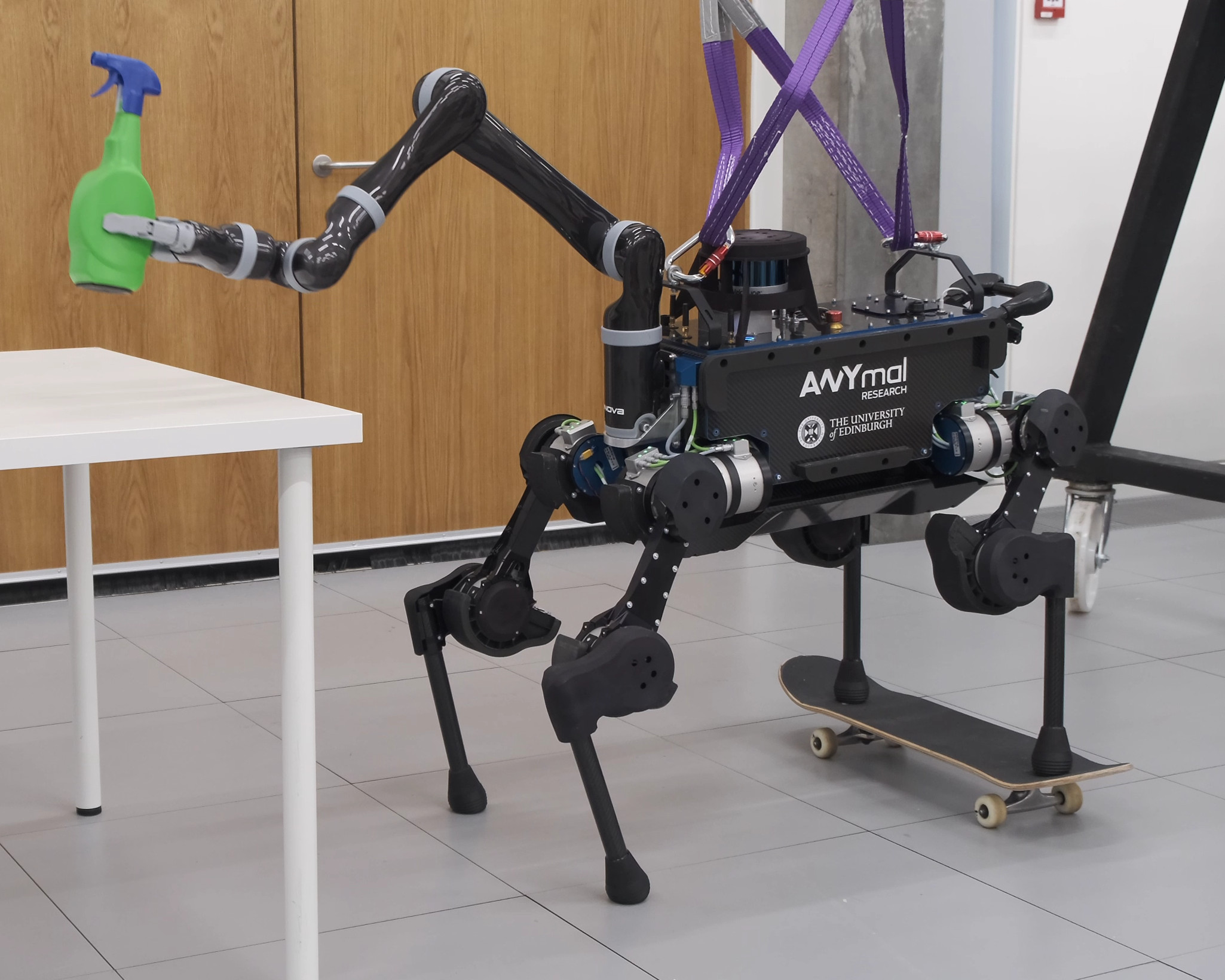}
        \caption{``Skateboard'' scenario}
        \label{subfigure:experiment_skateboard}
    \end{subfigure}
    \caption{Snapshots of the real robot executing the planned motions on a ramp (see \refFigure{subfigure:ramp}) and on a skateboard.}
    \label{figure:experiments}
\end{figure}

We executed the pick-and-place of a bottle on a table for different terrain: on flat ground and on a ``ramp'' (\refFigure{subfigure:experiment_ramp}).
The object being grasped was not modeled and it is therefore an external disturbance.
We also tested the trajectories for ground-feet friction coefficient mismatches by placing a skateboard underneath the feet of the robot (\refFigure{subfigure:experiment_skateboard}).
A video is available at \texttt{\url{https://youtu.be/vDesP7IpThw}}.

For the motions shown in the video, we optimized the trajectories at \SI{100}{\hertz} and then linearly interpolated them to \SI{400}{\hertz}.
It was the interpolation result that was then tracked by the controller.
We did this because computing optimal trajectories with $\mathcal{G}_3$ gets more computationally expensive as the problem discretization increases (see \refTable{table:times_convergence}).

\begin{figure}[ht]
    \centering
    \includegraphics[width=\linewidth]{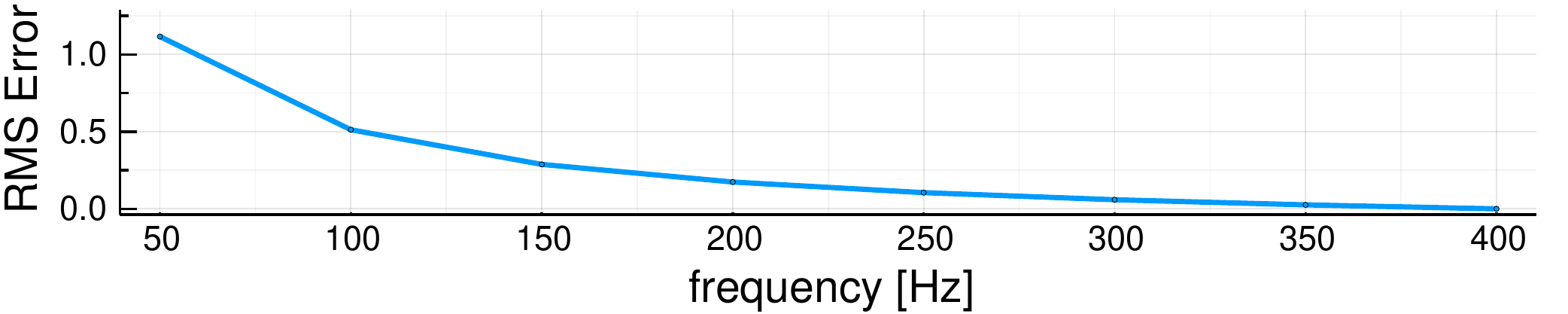}
    \caption{
        Root-mean-square error (RMSE) in newtons of the \gls{SUF} given different discretizations, for a \SI{400}{\hertz} baseline.
    }\label{figure:rms_error}
\end{figure}

To select the frequency of the trajectory before interpolation, we computed the root-mean-square error (RMSE) of the \gls{SUF} over time for the same trajectory using different discretization resolutions, with a \SI{400}{\hertz} resolution as a baseline.
As shown in \refFigure{figure:rms_error}, for a trajectory discretized at \SI{100}{\hertz} its \gls{SUF} $\mathrm{RMSE} \approx \SI{0.5}{\newton}$, which is acceptable for our purposes.

\section{Future Work}
\label{sec:future_work}

\begin{figure}[ht]
    \begin{subfigure}[t]{0.5\linewidth}
        \centering
        \includegraphics[width=0.6\linewidth]{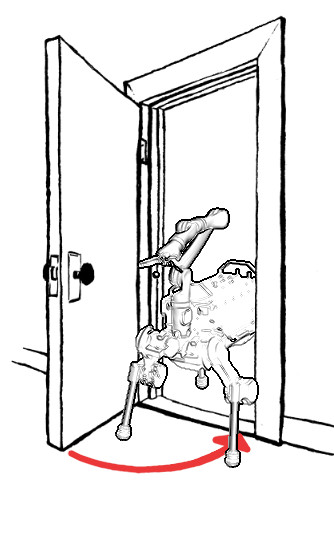}
    \end{subfigure}\hfill%
    \begin{subfigure}[t]{0.5\linewidth}
        \centering
        \includegraphics[width=0.6\linewidth]{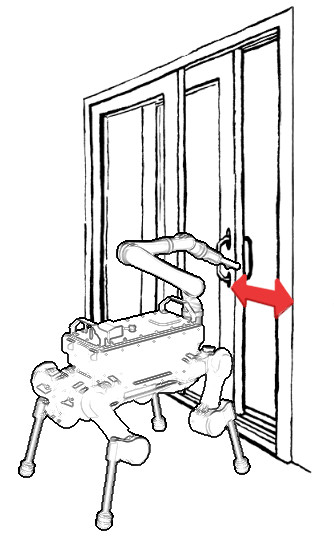}
    \end{subfigure}
    \caption{
        \textit{Left:} Door rotating over its hinges as a pivot point.
        \textit{Right:} Door sliding back and forth on a frame.
    }\label{figure:doors}
\end{figure}

\textit{Robustness Through Environment Exploitation:}
This work has focused on isotropic robustness, but for some tasks it may be more appropriate to be able to resist disturbances along particular directions, e.g., as in \refFigure{figure:doors}.
Based on our work \cite{ferrolho2019residual}, it should be straightforward to adapt \refSection{subsection:maximum_volume_inscribed_ball_of_a_polytopic_projection} to maximize either the volume of an ellipsoid or the magnitude of a specific vector inscribed in the projection of the dynamic force polytope.

\textit{Robust Dynamic Loco-manipulation:}
In this work, we did not study scenarios involving the making or breaking of support contacts with the environment.
It would be interesting to adapt the trajectory optimization transcription to allow for multi-contact tasks or for dynamic loco-manipulation.

\textit{Minimization of Robustness Loss:}
Our framework maximizes the robustness of a trajectory assuming reliable execution.
However, the system dynamics around a nominal trajectory are not linear, and given large enough perturbations the robot may be driven into areas of low robustness.
This can be alleviated by accounting for neighboring states and taking the control noise into account.

\section*{Acknowledgments}
We would like to thank Twan Koolen, Jo\~{a}o Moura, Romeo Orsolino, Fran\c{c}ois Pacaud, Theodoros Stouraitis, and Matt Timmons-Brown for their valuable help and feedback.
We would also like to thank the anonymous reviewers for their constructive comments.

\bibliographystyle{IEEEtran}
\bibliography{IEEEabrv,references}

\end{document}

%% file: acronyms.tex
\newacronym{AD}{AD}{automatic differentiation}
\newacronym{com}{CoM}{Center of Mass}
\newacronym{DDP}{DDP}{Differential Dynamic Programming}
\newacronym{DoF}{DoF}{Degrees of Freedom}
\newacronym{EoM}{EoM}{equation of motion}
\newacronym{FK}{FK}{forward kinematics}
\newacronym{GIWC}{GIWC}{Gravito-Inertial Wrench Cone}
\newacronym{IK}{IK}{inverse kinematics}
\newacronym{LP}{LP}{Linear Programming}
\newacronym{MPC}{MPC}{Model Predictive Control}
\newacronym{MRP}{MRP}{modified Rodrigues parameters}
\newacronym{NLP}{NLP}{nonlinear programming}
\newacronym{SUF}{SUF}{smallest unrejectable force}
\newacronym{TO}{TO}{trajectory optimization}
\newacronym{wrt}{w.r.t.}{with respect to}